\documentclass[10pt,twocolumn,letterpaper]{article}

\usepackage[pagenumbers]{wacv} 

\usepackage{xcolor}

\usepackage[T1]{fontenc}
\usepackage{amsmath,amssymb}
\usepackage{booktabs}
\usepackage{enumitem}
\usepackage{microtype}

\newcommand{\resti}{\textsc{ReSTI}}
\newcommand{\stibench}{\textsc{STI-Bench}}

\setlist{nosep,leftmargin=*}
\newcommand{\OriginalQACount}{2,064}
\newcommand{\ReSTIAccepted}{1,782}
\newcommand{\ReSTIExcluded}{282}

\newcommand{\LegacyMismatchCount}{1,014}
\newcommand{\SameAnswerCount}{652}
\newcommand{\ChangedAnswerCount}{547}
\newcommand{\RecontractedCount}{583}
\newcommand{\ExactDuplicateOptionRows}{32}
\newcommand{\NumericDuplicateOptionRows}{25}

\newcommand{\NoPoseTimeCount}{287}
\newcommand{\AmbiguousSubjectCount}{110}
\newcommand{\RandomFrameCount}{182}

\newcommand{\PotentiallyRecoverable}{182}
\newcommand{\StructurallyExcluded}{100}

\definecolor{wacvblue}{rgb}{0.21,0.49,0.74}
\usepackage[pagebackref,breaklinks,colorlinks,allcolors=wacvblue]{hyperref}

\title{\resti: A Source-Grounded Audit and Repair of \stibench{}}

\author{Pengzhan Sun\textsuperscript{1} \quad Ramanathan Rajaraman\textsuperscript{1} \quad Shiu-Hong Kao\textsuperscript{1}\\
Junbin Xiao\textsuperscript{2} \quad Angela Yao\textsuperscript{1}\\
\textsuperscript{1}National University of Singapore\\
\textsuperscript{2}University of Science and Technology of China\\
{\tt\small \{pengzhan,ayao\}@comp.nus.edu.sg}
}

\begin{document}
\raggedbottom
\maketitle
\begin{abstract}

Spatial--temporal benchmarks are valid only when their questions, source
annotations, and answer options identify the same physical quantity. We audit
\stibench{} against the official ScanNet, Waymo, and Omni6DPose sources and
find systematic coordinate-system and timestamp errors, under-specified targets and times, and disagreements between keyed options and answer details. We
introduce \resti{}, a source-backed revision that reconstructs every recoverable answer under an explicit target, time, coordinate system, physical quantity, and unit. 
Source reconstruction reveals task-level geometric failures: ScanNet Grounding omits the required alignment between annotation and raw camera coordinate systems, while Orientation measures camera rotation on the wrong plane. 
\resti{} replaces these labels with explicit, source-consistent
geometric definitions and corrects other source-verifiable defects, including Waymo poses evaluated at the wrong timestamp. 
Across \OriginalQACount{} legacy
questions, \resti{} retains \ReSTIAccepted{} questions and records \ReSTIExcluded{} evidence-backed exclusions. 
\resti{} therefore provides a conservative and source-traceable basis for evaluating precise video spatial--temporal reasoning.
Project page: \url{https://github.com/pengzhansun/ReSTI}.
\end{abstract}

\section{Introduction}

Spatial reasoning enables intelligent systems to understand where entities
are, how they relate to one another, and how they move through an
environment~\cite{liu2023vsr,jia2026omnispatial}.
In video, this capability extends beyond static scene layout to dynamic
quantities such as camera motion, object trajectories, distance, speed, and
orientation~\cite{li2025stibench,zhang2025dsibench}.
Benchmarks are the primary means of measuring progress on these
abilities, but their scores are meaningful only when questions and reference
answers faithfully encode the intended physical quantities. Benchmark
reliability is therefore a fundamental part of spatial-reasoning evaluation,
rather than merely a matter of dataset
cleanliness~\cite{northcutt2021pervasive,zhang2026revsi}.

\stibench{}~\cite{li2025stibench} is an ambitious test of precise
spatial--temporal reasoning over more than 300 videos from
ScanNet~\cite{dai2017scannet}, the Waymo Open Dataset~\cite{sun2020waymo}, and
Omni6DPose~\cite{zhang2024omni6dpose}. Its \OriginalQACount{} five-way
questions span eight tasks and broaden spatial evaluation from static metric
measurement and 3D grounding to dynamic quantities such as camera pose and
orientation changes, object displacement and path length, speed and
acceleration, and trajectory shape. This breadth makes \stibench{} a valuable
test bed, but it also makes correctness unusually sensitive to how source
annotations are converted into questions and answers. During reproduction, we
found that many released questions, source annotations, answer details, and
keyed choices could not all describe the same physical measurement.

\begin{figure*}[t]
  \centering
  \includegraphics[width=\textwidth]{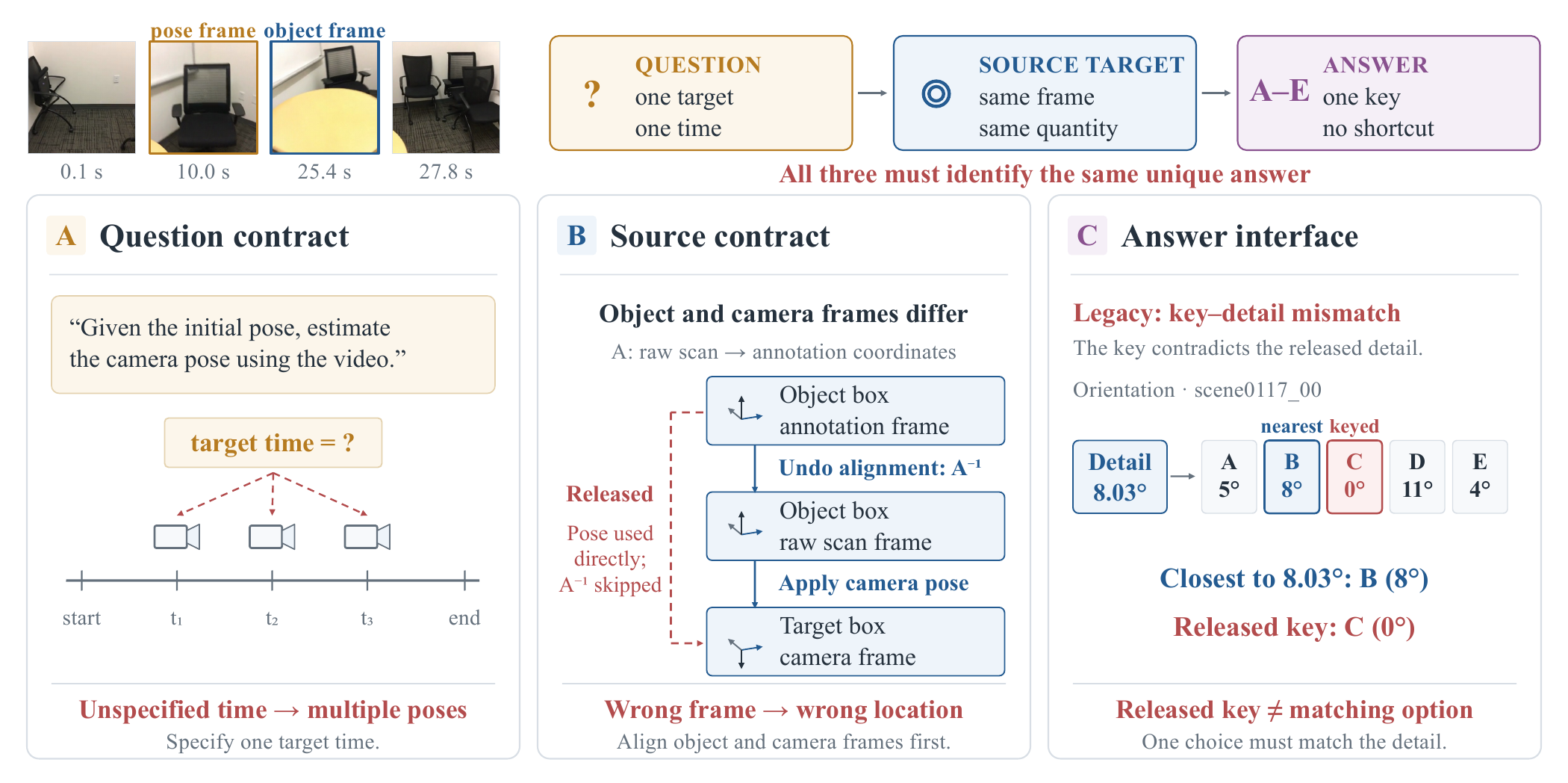}
  \caption{A valid benchmark item requires three mutually consistent layers.
  The question contract must specify a unique physical quantity; the source
  contract must reconstruct that quantity in the correct coordinate and time
  frames; and the answer interface must encode it without contradictory labels
  or option-only shortcuts. (A) The pose question omits its target time.
  (B) The released Grounding computation omits the conversion from annotation
  coordinates to the raw scan frame expected by the camera pose.
  (C) In a separate ScanNet example (scene0117\_00), the released detail is
  $8.03^\circ$ and option B is $8^\circ$, but the key selects option C
  ($0^\circ$).}
  \label{fig:teaser}
\end{figure*}

As shown in Figure~\ref{fig:teaser}, we organize benchmark reliability into three layers. At the
\emph{question layer}, each item must ask a meaningful, unambiguous question:
it must identify the target entity, target time, coordinate system, physical
quantity, and unit needed to determine one answer. At the \emph{source layer},
that answer must be deterministically derivable from the official source
annotations under explicit geometric and temporal conventions. At the
\emph{answer layer}, exactly one choice must agree with the derived reference;
the alternatives must represent distinct values. A failure at any one layer
contaminates the reported accuracy even if the other two layers are correct.
Making these conventions explicit follows the broader call to document
dataset provenance, construction, and intended use~\cite{gebru2021datasheets}.

We introduce \resti{}, a source-backed audit and revision of \stibench{}
organized around these three layers. At the question layer,
\NoPoseTimeCount{} pose questions omit the time at which a time-varying matrix
should be evaluated,
\AmbiguousSubjectCount{} motion questions name two possible subjects, and
\RandomFrameCount{} Grounding questions refer to an unnamed sampled video
time. These categories overlap, but each leaves the requested measurement
under-specified. \resti{} rewrites accepted questions to name the missing
target and time explicitly, and excludes an item when the source evidence
cannot resolve its ambiguity.

At the source layer, reconstruction exposes systematic geometric and temporal
errors rather than isolated transcription noise. ScanNet Grounding combines
annotation-world object boxes with raw-world camera poses without the required
alignment transform. Orientation computes a camera-axis rotation on the wrong
coordinate plane instead of the intended horizontal heading. For Waymo Pose
Estimation, source reconstruction shows that 90 of 121 retained legacy
matrices were taken from the wrong timestamp.
\resti{} replaces these operations with explicit, source-consistent
definitions, yielding \ChangedAnswerCount{} corrected source-backed values and
\RecontractedCount{} values re-derived under explicit contracts.

At the answer layer, the released key disagrees with its answer detail in
\LegacyMismatchCount{} rows; \ExactDuplicateOptionRows{} rows contain exact
duplicate choices, and \NumericDuplicateOptionRows{} contain textually
different choices with the same numeric value. \resti{} rebuilds keys and
alternatives and checks both value uniqueness and key--detail agreement.
We also test whether choices reveal the answer without video. This follows
the general lesson from annotation-artifact studies: construction cues can
support prediction without the intended
evidence~\cite{gururangan2018artifacts,geirhos2020shortcut}.
In visual question answering, complementary-image balancing and evaluation
under changed answer priors similarly test whether predictions depend on
visual evidence~\cite{goyal2017making,agrawal2018priors}.

Overall, \resti{} retains \ReSTIAccepted{} source-backed questions and records
\ReSTIExcluded{} evidence-backed exclusions, accounting for all
\OriginalQACount{} legacy rows. Among the retained questions,
\SameAnswerCount{} preserve the legacy answer value, \ChangedAnswerCount{}
receive a corrected value, and \RecontractedCount{} are re-derived under an
explicit new contract. The resulting release provides a conservative,
source-traceable basis for evaluating precise spatial--temporal reasoning and
for distinguishing model errors from benchmark-construction errors.

\section{Related Work}

\begin{figure*}[t]
  \centering
  \includegraphics[width=\textwidth]{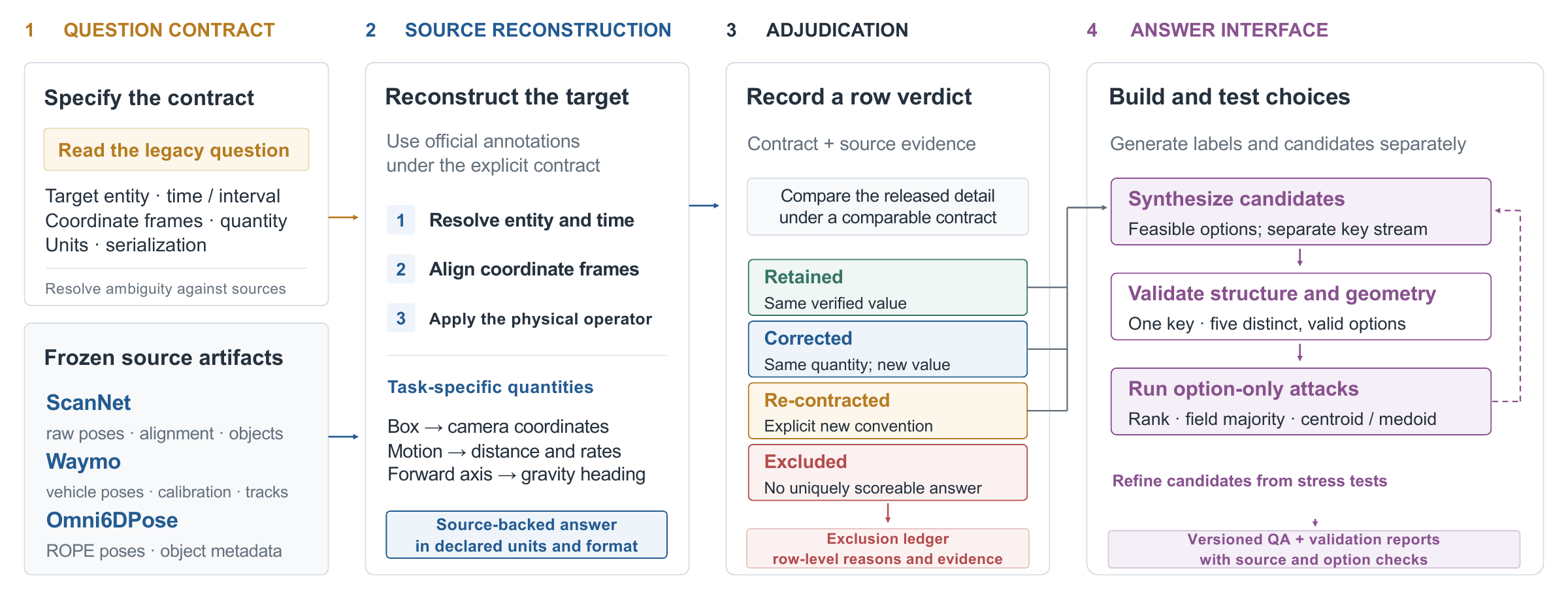}
  \caption{The \resti{} audit and repair pipeline. (1) Specify the question
  contract and identify the source annotations. (2) Resolve the entity and
  time, align coordinate frames, and apply the physical operator to reconstruct
  the target. (3) Compare the released answer with the reconstructed target
  under a comparable contract and record a retained, corrected, re-contracted,
  or excluded verdict. (4) Synthesize candidates, validate their structure and
  geometry, and test for option-only shortcuts. Exclusions carry reasons and
  evidence; option-only tests guide candidate refinement.}
  \label{fig:repair-pipeline}
\end{figure*}

Spatial-reasoning evaluation builds on compositional visual question
answering: CLEVR~\cite{johnson2017clevr} provides controlled diagnostic
scenes, while GQA~\cite{hudson2019gqa} derives questions from real-image
scene graphs and controls answer distributions. Benchmarks now span image-based
relations~\cite{liu2023vsr,kamath2023whatsup}, visual spatial
cognition~\cite{fu2024blink,jia2026omnispatial}, and video-based understanding
of scenes and motion~\cite{yang2025thinking,zhang2025dsibench,lin2025mmsivideo}.
VSR~\cite{liu2023vsr} and What'sUp~\cite{kamath2023whatsup} test spatial
relations between objects, with What'sUp using controlled photographs that
vary the relation while preserving object identities.
SpatialSense~\cite{yang2019spatialsense} uses adversarial crowdsourcing to
reduce reliance on language priors and simple 2D spatial cues.
BLINK~\cite{fu2024blink} adds depth, correspondence, and multi-view perception
tasks using single or multiple images, while OmniSpatial~\cite{jia2026omnispatial}
covers spatial logic, interaction, dynamic reasoning, and perspective-taking.
In reconstructed 3D scenes, ScanRefer~\cite{chen2020scanrefer} localizes
objects from descriptions, and ReferIt3D~\cite{achlioptas2020referit3d}
distinguishes target instances using referential language.
ScanQA~\cite{azuma2022scanqa} extends object grounding to scene question
answering, while SQA3D~\cite{ma2023sqa3d} conditions questions on an
agent's described position and orientation.
At the video level,
VSI-Bench~\cite{yang2025thinking} evaluates scene configuration, metric
measurement, and appearance order from egocentric videos.
DSI-Bench~\cite{zhang2025dsibench} examines coupled observer and object motion;
OSI-Bench~\cite{wu2026osibench} extends metric and kinematic reasoning to
open-world pedestrian videos with sensor-derived geometry; and
MMSI-Video-Bench~\cite{lin2025mmsivideo} spans perception, planning,
prediction, and cross-video reasoning.
UCS-Bench~\cite{wang2026ucsbench} evaluates continual
spatial reasoning in egocentric video streams, requiring memory of object
locations relative to a moving user.
Within this landscape, \stibench{}~\cite{li2025stibench} combines static spatial
tasks with explicit dynamic measurements of displacement, path length, speed,
acceleration, camera orientation, trajectories, and pose.
This broader coverage makes benchmark reliability increasingly dependent on
the annotations and procedures used to derive answers. General test-set
audits show that label errors can alter model comparisons~\cite{northcutt2021pervasive};
within video spatial evaluation, ReVSI~\cite{zhang2026revsi} revisits
VSI-Bench by correcting object and geometry annotations, regenerating verified
questions, and aligning evaluation with the visual evidence available under
different frame budgets. Complementing these efforts, \resti{} diagnoses
systematic defects in \stibench{}, including under-specified questions,
coordinate-frame and timestamp errors, and contradictory or shortcut-prone
answer options. We reconstruct recoverable targets from official source
annotations under explicit measurement contracts, correct source-verifiable
errors, and document evidence-backed exclusions.

\section{Audit and Repair Protocol}
\label{sec:protocol}

\begin{figure*}[t]
  \centering
  \includegraphics[width=\textwidth]{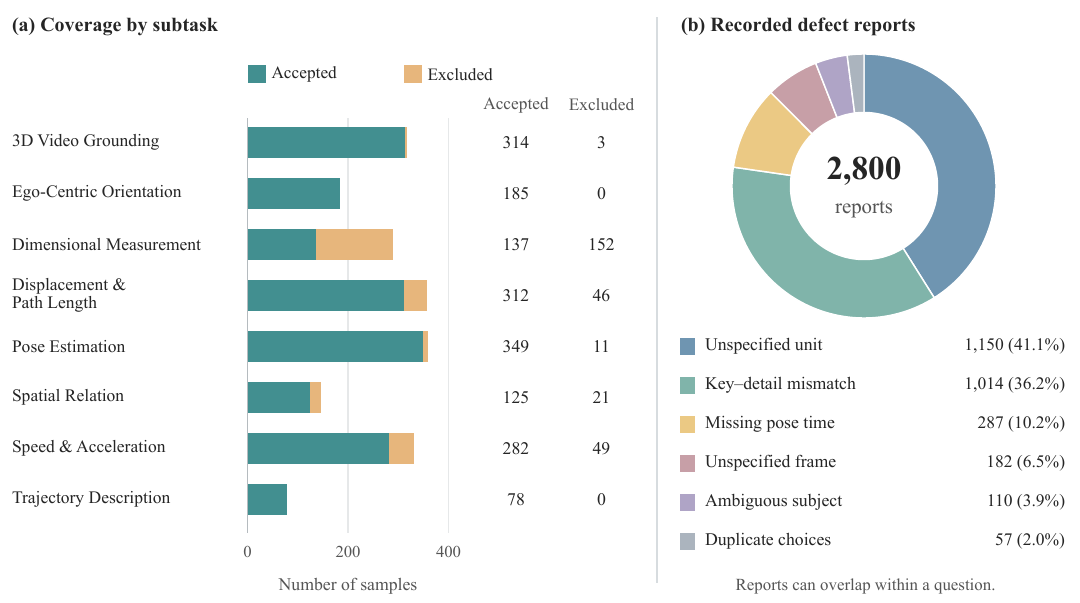}
  \caption{\textbf{Audit overview.} \textbf{(a)} Accepted and excluded questions
  by subtask, with exact counts. \resti{} accepts
  \ReSTIAccepted{} of \OriginalQACount{} questions; accepted includes
  retained, corrected, and re-contracted items. \textbf{(b)} Distribution of
  2,800 defect reports across six categories. Percentages use the sum of these
  reports, not the number of questions: one question can contribute to multiple
  categories. }
  \label{fig:audit-overview}
\end{figure*}

Our protocol traces each benchmark answer from its written question to the
source annotations and selectable options. Figure~\ref{fig:repair-pipeline}
organizes this process into question specification, source reconstruction,
adjudication, and answer-interface checks.

\subsection{Question contract}

Measurement validity requires making explicit how a claimed capability is
operationalized by observable quantities~\cite{jacobs2021measurement}.
For this audit, we determine whether a question has one verifiable answer by
representing it with
an explicit measurement contract
\begin{equation}
  \mathcal{C}=(s,e,t,\mathcal{F},q,u,\sigma,\mathcal{D}),
  \label{eq:contract}
\end{equation}
where $s$ identifies the source annotations, $e$ the target entity, $t$ the time
or interval, $\mathcal{F}$ the relevant coordinate frames, $q$ the physical
operator, $u$ the unit, $\sigma$ the answer-formatting rule, and $\mathcal{D}$
the distractor policy. We read the released question to identify
$(e,t,\mathcal{F},q,u)$ and resolve missing specifications against the source
evidence where possible. A valid item requires an unambiguous question, a
source-consistent target, and exactly one matching option without a tested
option-only shortcut. These three requirements distinguish question ambiguity,
source derivation errors, and answer-interface defects.

\subsection{Source reconstruction}

We reconstruct the requested quantity from official source annotations under
the explicit contract. For ScanNet, we use camera poses, scene alignment, and
object annotations; for Waymo, timestamped vehicle poses, camera calibration,
3D boxes, and object tracks; and for Omni6DPose, object poses and metadata.
Following the source reconstruction panel in Figure~\ref{fig:repair-pipeline},
we first resolve the target entity and time, then align the relevant coordinate
frames, and finally apply the physical operator. This yields camera-relative
boxes for Grounding, distances and rates from motion, or gravity-relative
heading changes for Orientation, expressed in the declared units and format.
We record the source evidence and derivation for each question so that the
reconstructed target can be traced to its inputs. This process checks the use
of existing source annotations; it does not re-annotate the source datasets.

\subsection{Adjudication}

Adjudication determines whether the released answer can be verified or
repaired under the available contract and source evidence. As shown in
Figure~\ref{fig:repair-pipeline}, we compare the released answer detail with
the reconstructed target when both refer to the same physical quantity under
comparable conventions. Each question receives one of four verdicts:
\emph{Retained} preserves the same verified value; \emph{Corrected} replaces
an incorrect value for the same quantity; \emph{Re-contracted} derives a value
under an explicit convention when the original convention is invalid,
unstated, or incomparable; and \emph{Excluded} records that no uniquely
scoreable answer can be established. The first three outcomes proceed to
answer construction. Excluded questions enter a separate record with reasons
and supporting evidence. This distinction prevents a newly defined target
from being counted as a correction to a comparable legacy value.

\subsection{Answer interface}

The answer interface turns each accepted target into five choices and checks
whether the choices preserve the intended reasoning task. Following the final
panel of Figure~\ref{fig:repair-pipeline}, we first \emph{synthesize candidates}
subject to task-specific feasibility and separation constraints, with answer
letters assigned independently of candidate generation. We then
\emph{validate structure and geometry}: each item must have five distinct,
valid options and exactly one keyed option matching the source-derived answer.
Checks include finite values, units, timestamp bounds, and valid pose and box
rotations. Finally, we \emph{run option-only attacks}, including numeric rank,
field majority, and proximity to the candidate centroid or medoid, to detect
answers recoverable without video or question content. These tests complement
label verification because geometrically correct choices can still reveal the
key. Adversarial filtering in SWAG~\cite{zellers2018swag} provides a
related precedent for testing candidate-construction artifacts; our checks
target the numeric and geometric structure of spatial answers.
The final Orientation choices replace the symmetric grid with jittered
angular alternatives and balanced signed-value ranks, preserving the validated
heading targets (Section~\ref{sec:orientation}).

\section{STI-Bench Diagnosis}
\label{sec:diagnosis}

The audit finds defects at every layer of the measurement contract.
Figure~\ref{fig:audit-overview}(b) summarizes the relative frequency of six common
construction defects in the released questions and options; source-level geometric and temporal failures
are quantified separately below.

\paragraph{The option interface often contradicts the reference.}
Among the \LegacyMismatchCount{} key--detail mismatches, 315 involve structured
Grounding or Pose records, 173 are scalar discrepancies not explained by
rounding, and the remainder include incompatible formats or coarser numeric
rounding. The duplicate-candidate total combines
\ExactDuplicateOptionRows{} reports of exact duplicate strings and
\NumericDuplicateOptionRows{} reports of textually different choices that
encode the same numeric value. These are interface failures even before any
source dataset is consulted.

\paragraph{Many questions do not select one measurement.}
The omitted times, unnamed frames, ambiguous subjects, and unit tags in
Figure~\ref{fig:audit-overview}(b) leave components of Eq.~\ref{eq:contract} unspecified.
Additional cases ask about grouped objects, give zero-length intervals, or
print the answer in the premise. Such rows cannot be repaired by choosing the
candidate nearest to the legacy detail because the written task itself does
not identify one ground truth.

\paragraph{Source reconstruction exposes systematic numeric failures.}
ScanNet Grounding combines object and camera quantities from different world
frames (Section~\ref{sec:grounding}); Orientation uses a source-axis component
rather than a gravity-relative heading (Section~\ref{sec:orientation}); and 90
of 121 retained legacy Waymo pose matrices were taken from the wrong timestamp.
Replacing them with source poses at the explicitly stated target time changes
translation by a median of 0.77\,m and a maximum of 31.6\,m. These patterns are
too structured to treat as isolated rounding or transcription errors.

\paragraph{The choices reveal answers without video.}
On 1,129 legacy scalar rows, selecting the median numeric candidate scores
46.9\%, compared with 20\% chance. The best constant letter reaches 23.8\%, and
the correct value is the smallest candidate for all 81 Waymo Orientation rows.
These attacks do not establish that a video model used the same shortcut, but
they show that the benchmark permits accuracy unrelated to its intended visual
evidence.

\section{Systematic Errors in 3D Video Grounding}
\label{sec:grounding}

\begin{figure*}[t]
  \centering
  \includegraphics[width=\textwidth]{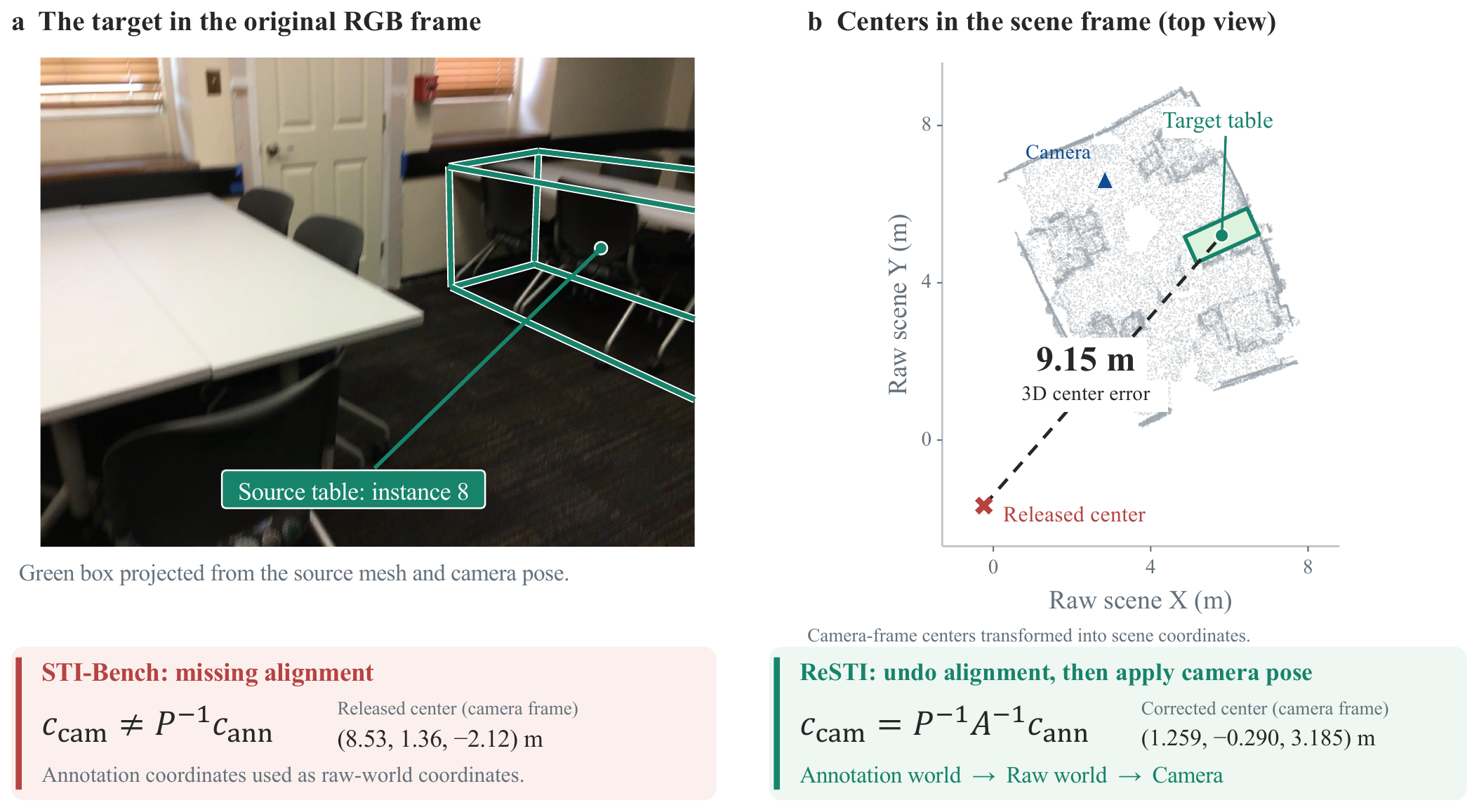}
  \caption{\textbf{A Grounding label displaced from its target by 9.15\,m.}
  \textbf{(a)} Original ScanNet RGB frame with the source table box projected
  using its mesh annotation, axis alignment, camera pose, and intrinsics.
  \textbf{(b)} Top-down view of the source mesh. Green marks the table; the red
  cross is the signed \texttt{Answer Detail} center mapped back into the raw
  scene. The dashed line connects their centers; the distance is measured in
  3D. The lower panels contrast the invalid frame substitution with the
  correct transform. Their reported tuples are camera-frame centers; panel
  (b) displays both centers after transformation into the common scene frame.
  This displacement is independent of how the legacy scalar heading is interpreted.}
  \label{fig:grounding}
\end{figure*}

\subsection{Diagnosis: object and camera in different worlds}

ScanNet Grounding exhibits a systematic coordinate-frame mismatch in the
derivation of camera-relative object boxes. The target is an oriented object
box expressed in the camera frame at a specified time, but the legacy values
combine two incompatible world frames. Let
\begin{align*}
  \mathbf{A} &= \mathbf{T}_{\mathrm{ann}\leftarrow\mathrm{raw}}, &
  \mathbf{P}(t) &= \mathbf{T}_{\mathrm{raw}\leftarrow\mathrm{cam}}(t),
\end{align*}
where $\mathbf{A}$ is the ScanNet axis-alignment transform and $\mathbf{P}(t)$
is the raw camera pose. For an annotation-world center
$\mathbf{c}_{\mathrm{ann}}$, the camera-space center is
\begin{equation}
  \mathbf{c}_{\mathrm{cam}}(t)
  =\mathbf{P}(t)^{-1}\mathbf{A}^{-1}\mathbf{c}_{\mathrm{ann}}.
  \label{eq:grounding-center}
\end{equation}
Here rigid transforms act on points in homogeneous coordinates. The released
centers are instead consistent with applying
$\mathbf{P}(t)^{-1}$ directly to $\mathbf{c}_{\mathrm{ann}}$, which supplies
an annotation-world point to a raw-world transform.

An independently built source-native object trace distinguishes this hypothesis
from ordinary matching noise. The trace does not use the legacy camera-frame
box. Corrected centers agree with this reconstructed physical geometry
to a median 0.006\,m, whereas released centers lie a median 2.61\,m from the
nearest physical object. In the example of Figure~\ref{fig:grounding},
the target is a white table in ScanNet scene0041\_00 at 4.0\,s.
Back-projecting the signed center in the released \texttt{Answer Detail}
places it outside the reconstruction, 9.15\,m from the source table center.
It lies only 0.004\,m from the table's annotation-world numeric center,
consistent with omitting the alignment conversion. This residual measures
agreement with the faulty computation, not repaired localization accuracy.
The displacement alone establishes a center error, independently of how the
released scalar heading is interpreted.

\subsection{Repair: transform a typed oriented box}

\resti{} transforms the complete box only after camera and object are placed
in a common world:
\begin{equation}
  \mathbf{T}_{\mathrm{cam}\leftarrow\mathrm{box}}(t)
  =\mathbf{P}(t)^{-1}\mathbf{A}^{-1}
   \mathbf{T}_{\mathrm{ann}\leftarrow\mathrm{box}}.
  \label{eq:grounding-box}
\end{equation}
The same typed output contract is implemented source by source. Waymo boxes
pass through timestamped vehicle pose and calibrated camera extrinsics;
Omni6DPose ROPE poses are normalized under its verified frame conversion. All
answers serialize box-local edge lengths, signed camera-frame centers, and
complete box-to-camera rotations in OpenCV axes ($+X$ right, $+Y$ down,
$+Z$ forward). A scalar heading is retained only as a redundant summary.

This reconstruction retains 314 of 317 Grounding rows: 179 ScanNet, 80 Waymo,
and 55 Omni6DPose. Three ScanNet rows are excluded because the description does
not identify one source object. Complete rotations, not only centers and
headings, are checked for finiteness, orthonormality, and consistent
serialization.

We also rebuild the choices by jointly sampling alternatives in center,
log-dimensions, and rotation, with source-scaled separation constraints and
independently assigned answer positions. This replaces an earlier field-wise
generator whose per-field majority revealed the answer on all 314 rows.
The strongest tested option-only attack with full coverage scores 22.61\%
after repair, compared with 20\% five-way chance.

\section{Systematic Errors in \mbox{Ego-Centric Orientation}}
\label{sec:orientation}

\begin{figure*}[t]
  \centering
  \includegraphics[width=\textwidth]{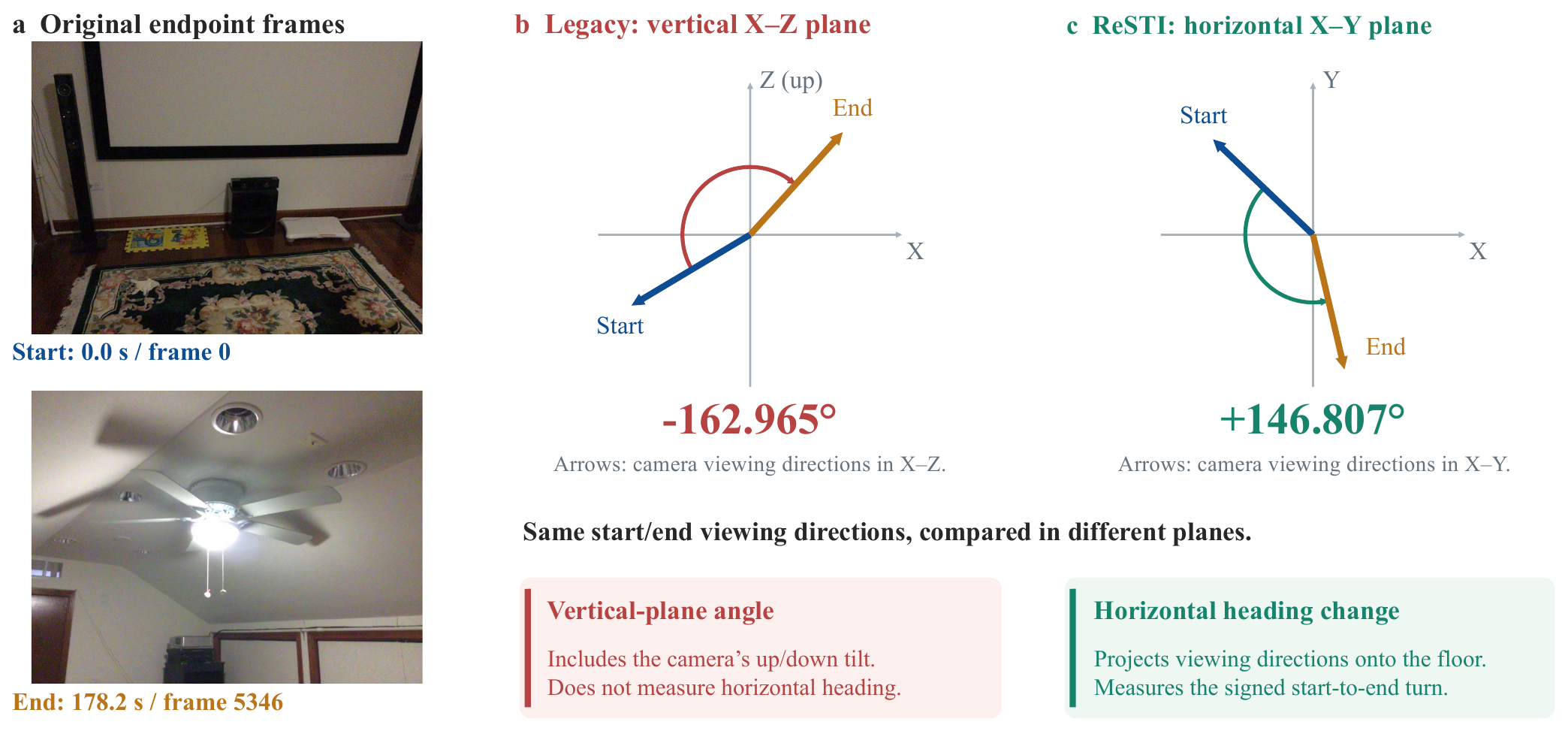}
  \caption{\textbf{A vertical-plane angle is not horizontal heading change.}
  \textbf{(a)} Original RGB endpoints of ScanNet scene0012\_00; times follow
  the STI video frame indexing. \textbf{(b)} Normalized camera-forward vectors
  projected onto raw-world X--Z, a vertical plane, reproduce the released
  detail. \textbf{(c)} Projection onto the gravity-horizontal X--Y plane gives
  the repaired heading. Blue and gold denote the start and end directions;
  arcs show signed angles from the start direction to the end direction.
  The vertical-plane angle includes up/down tilt, whereas the horizontal
  projection measures heading change on the floor plane. Both calculations
  use the same original source poses.}
  \label{fig:orientation-case}
\end{figure*}

\subsection{Diagnosis: a source-axis component is not heading}

Ego-Centric Orientation exhibits a systematic definition error: the task asks
for the camera's orientation change, but the legacy values are consistent with
an operator tied to arbitrary source axes. For a ScanNet camera-to-world
rotation $\mathbf{R}_{W\leftarrow C}(t)$, the released detail is recovered by
selecting one camera-axis column, projecting it onto raw-world $X$--$Z$, and
measuring a signed endpoint angle. ScanNet $Z$ is vertical, so this projection
describes a vertical-plane component rather than gravity-horizontal heading.
The same column pattern appears in Waymo even though that column represents a
different physical camera axis. A vertical-plane angle includes camera tilt
and does not measure the horizontal heading change.

Figure~\ref{fig:orientation-case} contrasts the two projection planes for
ScanNet scene0012\_00. From the original poses at frames 0 and 5346, the
raw-world $X$--$Z$ computation gives $-162.965^{\circ}$, matching the
released detail of $-162.97^{\circ}$. Projecting the same start and end
optical-forward vectors onto the horizontal $X$--$Y$ plane gives
$+146.807^{\circ}$, which rounds to the \resti{} answer
$+146.81^{\circ}$. The difference comes from the projection plane;
both calculations use the same video endpoints and source poses.

The source comparison supports a definition failure rather than random pose
noise. An independently implemented gravity-relative operator reproduces all
102 corrected targets in the checked subset. By contrast, the best of five
standard heading conventions reproduces the legacy target in only 14 of 104
ScanNet cases. These counts validate the replacement operator; they do not
claim access to the private code that generated the legacy labels.

\subsection{Repair: gravity-relative optical heading}

Let $\mathbf{a}$ be the optical-forward axis in the source camera convention,
$\mathbf{u}$ the world-up unit vector, and $\mathbf{R}(t)$ the camera-to-world
rotation. We first project optical forward onto the plane orthogonal to gravity,
then measure the signed endpoint angle:
\begin{align}
  \mathbf{f}(t) &= \mathbf{R}(t)\mathbf{a}, \\
  \mathbf{h}(t) &=
  \frac{\mathbf{f}(t)-\mathbf{u}\bigl(\mathbf{u}^{\top}\mathbf{f}(t)\bigr)}
       {\left\|\mathbf{f}(t)-\mathbf{u}
       \bigl(\mathbf{u}^{\top}\mathbf{f}(t)\bigr)\right\|}, \\
  \theta &= \operatorname{atan2}\!\left(
    \mathbf{u}^{\top}(\mathbf{h}_0\!\times\!\mathbf{h}_1),
    \mathbf{h}_0^{\top}\mathbf{h}_1\right).
  \label{eq:heading}
\end{align}
The signed result is serialized in degrees in $[-180,180)$. ScanNet uses camera
$+Z$ as optical forward and raw-world $+Z$ as up; Waymo uses camera $+X$ as
forward and ENU $+Z$ as up. Applying the same rigid rotation to the camera
trajectory and world basis leaves Eq.~\ref{eq:heading} unchanged; the maximum
measured invariance error is $5.69\times10^{-14}$ degrees.

All 185 Orientation rows are re-contracted under this definition. Nine ScanNet
sequences end with invalid raw poses, so their intervals are shortened to the
last valid timestamp and that endpoint is stated in the question. No invalid
nominal pose is silently replaced.

\subsection{Repair: regenerated Orientation choices}

The final release also repairs the Orientation answer choices. An earlier
revision placed four distractors at $\pm7.5^{\circ}$ and $\pm15^{\circ}$ around
the correct angle. This symmetric grid exposed the answer: selecting the
candidate nearest the circular mean solved all 185 questions without video,
and selecting the median solved 179. Correcting the heading labels therefore
also required rebuilding the choices.

We regenerate all 185 option sets using jittered angular offsets, wrapped to
$[-180,180)$ with at least $1^{\circ}$ pairwise angular separation. The
construction balances the correct answer's signed-value rank within each
source and assigns answer letters independently. Every question and
source-derived answer value is preserved. Across all 185 questions, median-pick
accuracy falls from 96.8\% to 20.0\%, and the best fixed signed-value rank scores
20.5\%. This repair removes the symmetric-grid construction leak while
preserving the validated gravity-relative heading targets.

\section{Task Coverage and Additional Repairs}
\label{sec:remaining}

Figure~\ref{fig:audit-overview}(a) summarizes acceptance and exclusion across all eight
\resti{} subtasks, including Grounding and Orientation from
Sections~\ref{sec:grounding} and~\ref{sec:orientation}. The other six tasks do
not share one geometric root cause, so we apply the same contract separately
to each source--task pair. Exclusion means that the available evidence does
not identify one scoreable answer; it does not necessarily mean the legacy
value is numerically false. The paragraphs below explain their largest repairs.

\paragraph{Measurements require identifiable referents.}
Dimensional Measurement derives one specified dimension of a matched source
object. Spatial Relation preserves 98 verified mounted-camera relations and
uses source-identifiable Omni6DPose referents. For Displacement and Path
Length, we recompute endpoint displacement and integrated path from source
motion under named subjects and intervals.

\paragraph{Rate questions require a named statistic.}
The 115 ScanNet questions described as ``average speed'' use endpoint
displacement divided by elapsed time, the magnitude of average velocity, rather
than path length divided by time. \resti{} states that statistic explicitly.
Seventy-eight values remain valid under the clarified contract and 37 change
after source verification. This repair preserves the recoverable quantity
without presenting two physically different rates under one name.

\paragraph{Pose matrices require one frame and one time.}
Rewritten pose questions state a target timestamp, world frame, camera axes,
and matrix direction. Among 121 retained Waymo pose rows, 90 legacy matrices
were taken from the wrong source timestamp. Replacing them with the source pose
at the stated target time changes translation by a median of 0.77\,m and a
maximum of 31.6\,m. In contrast, all 144 retained ScanNet pose values remain
unchanged: their labels were recoverable, but the original questions omitted
time and their distractors reused identical rotations. This comparison shows
why wording, target reconstruction, and options require separate verdicts.

\paragraph{Trajectory descriptions require continuous source motion.}
All 78 Waymo trajectory answers are rebuilt from timestamped motion and turn
segmentation. Legacy distances frequently lie on a 10\,m grid (for example,
340\,m for a 346.8\,m source path), which is too coarse to support the precision
implied elsewhere in the task. The repair retains the qualitative trajectory
structure while serializing source-derived distances.

\paragraph{Exclusions are evidence-bearing outcomes.}
The \ReSTIExcluded{} excluded rows carry row-level reasons. The largest groups
contain multiple objects matching the description, grouped referents, missing
annotated counterparts, or no visible object reproducing the premise. Other
reasons include out-of-frustum targets, zero-length intervals, premise--answer
identity, two possible motion subjects, no dominant relation, and undefined
dimensions. Of these exclusions, \PotentiallyRecoverable{} concern ambiguous
or unmatched referents, and \StructurallyExcluded{} concern defective task
statements.

\section{Limitations}

\paragraph{Dependence on source annotations.}
The main limitation of \resti{} is that its reference answers depend on the
quality of the existing source annotations. We reconstruct benchmark targets
from the official ScanNet, Waymo, and Omni6DPose annotations, but do not
re-annotate those datasets to independently establish their accuracy or
completeness. Missing objects, inaccurate boxes, or errors in source poses and
tracks can therefore propagate into the reconstructed answers. This
dependence reflects the broader problem of data-quality issues propagating
through later processing stages~\cite{sambasivan2021cascades}. Agreement with
the source establishes consistency with that evidence, but cannot guarantee
that every answer matches the physical scene. Our audit addresses how
\stibench{} specifies questions and derives answers from its sources.
Source re-annotation is outside the scope of this audit.

\paragraph{Scope of the evaluation.}
This report evaluates benchmark construction through source reconstruction,
geometric consistency, and answer-interface checks. These results establish
properties of the questions and reference answers; they do not measure changes
in model accuracy or rankings.

\section{Conclusion}

We present a diagnosis and repair pipeline that traces spatial--temporal
benchmark questions through source reconstruction, adjudication, and
answer-interface checks. Applied to \stibench{}, the pipeline identifies
systematic annotation and answer-construction errors, particularly the
coordinate-frame mismatch in 3D Video Grounding and the incorrect rotation
plane in Ego-Centric Orientation. We rebuild the benchmark as \resti{},
retaining \ReSTIAccepted{} questions with source-backed targets and recording
\ReSTIExcluded{} evidence-backed exclusions. By making the requested quantities
explicit and correcting their derivations, \resti{} provides a clearer basis
for separating model reasoning errors from benchmark-construction errors.
Its reliability remains bounded by source-annotation quality.

{
    \small
    \bibliographystyle{ieeenat_fullname}
    \bibliography{main}
}

\end{document}